\documentclass[letterpaper]{article} 
\usepackage{aaai23}  
\usepackage{times}  
\usepackage{helvet}  
\usepackage{courier}  
\usepackage[hyphens]{url}  
\usepackage{graphicx} 
\usepackage{natbib}  
\usepackage{caption} 
\usepackage{algorithm}
\usepackage{dirtytalk}
\usepackage{algpseudocode}
\usepackage{multirow}
\usepackage{amsmath}
\usepackage{subcaption}
\usepackage{newfloat}
\usepackage{listings}
\DeclareCaptionStyle{ruled}{labelfont=normalfont,labelsep=colon,strut=off} 
\floatstyle{ruled}
\newfloat{listing}{tb}{lst}{}
\floatname{listing}{Listing}
\title{\say{If you build they will come}: Automatic Identification of News-Stakeholders to detect Party Preference in News Coverage}
\author {
    Alapan Kuila,
    Sudeshna Sarkar
}
\affiliations {
    IIT Kharagpur\\
    alapan.cse@iitkgp.ac.in,
    sudeshna@cse.iitkgp.ac.in,
}

\usepackage{bibentry}

\begin{document}

\maketitle

\begin{abstract}
The coverage of different stakeholders mentioned in the news articles significantly impacts the slant or polarity detection of the concerned news publishers. For instance, the pro-government media outlets would give more coverage to the government stakeholders to increase their accessibility to the news audiences. In contrast, the anti-government news agencies would focus more on the views of the opponent stakeholders to inform the readers about the shortcomings of government policies. In this paper, we address the problem of stakeholder extraction from news articles and thereby determine the inherent bias present in news reporting. Identifying potential stakeholders in multi-topic news scenarios is challenging because each news topic has different stakeholders. The research presented in this paper utilizes both contextual information and external knowledge to identify the topic-specific stakeholders from news articles. We also apply a sequential incremental clustering algorithm to group the entities with similar stakeholder types. We carry out all our experiments on news articles on four Indian government policies published by numerous national and international news agencies. We also further generalize our system, and the experimental results show that the proposed model can be extended to other news topics.
\end{abstract}

\section{Introduction}

One of the primary responsibilities of the mass media is to inform the public about different socio-political and eco-cultural issues and their effects on day to day life. The news articles depict these topics and relevant stakeholders from different perspectives and help citizens to build their own perceptions. However, any conflicting issue contains multiple stakeholders with their contradicting viewpoints. In an ideal case, journalists should cover all those viewpoints and should give balanced saliency to all the participating stakeholders in their reporting. The coverage of political actors in media coverage is important as it will increase their accessibility to the audiences, influencing their political judgements.

Each news report consists of factual information on relevant topics along with a number of named entities of different types e.g. person, place, geopolitical entity, organization, date, time etc. Some of these entities (mainly person, organization or geo-political entities) are directly involved in the news story as the main actors of the conflict or as the opinion holder of some specific claims.  These opinions or claims are subjective statements made by the actors about the political, social, economical issues or other actors expressing criticism, support or enmity. These actors are referred to as stakeholders. For instance, in news reporting on the American presidential election, two major political parties and their presidential candidates are the potential stakeholders(~\cite{d2000media}). In the Indian general election where multiple parties participate, the news articles consider all the participating political parties, their electoral candidates, election commission and last but not the least, the voters as the prospective stakeholders. 

Stakeholder identification is a critical task in news analysis with several useful applications. Firstly, extraction of all the potential stakeholder mentions assists in determining salient named entity extraction; secondly, the extraction of stakeholders with their corresponding opinions and context sometimes helps in political ideology detection; thirdly, the presence of the same stakeholder entities in two different news reports helps to find relatedness between news-events. The imbalance in visibility or coverage of these stakeholders helps in the quantitative analysis of news bias. We consider a stakeholder to be visible or covered in a media publication if at least one candidate of that stakeholder type is mentioned in the published news stories. We use the term \textit{coverage} and \textit{visibility} interchangeably throughout this paper.


However, each news topic consists of a different set of stakeholder types. For example, in the case of football-related news, football club management, players, coaches, and ex-players are the probable stakeholders, whereas, in the case of entertainment news, actors, directors, critics, and movie-viewers are the prospective stakeholders. Therefore, creating a well-defined tag set or annotated data for stakeholder classification in a multi-topic news scenario is a cumbersome task. Moreover, the absence of a well-defined tag set and scarcity of annotated data make the problem more challenging. In these circumstances, designing a generalized system for stakeholders' identification is challenging in the news domain. Existing research works rely on the dictionary-based matching approach, where they look up the candidate entities in the build-in stakeholder-specific dictionaries to identify the stakeholder types. However, the problem would appear when a stakeholder candidate is not present in those dictionaries. With the upswing of modern NLP techniques, Some researchers also try to identify the bias by choice of words for referencing the named entities in media reports (~\cite{Hamborg2019AutomatedIO};~\cite{spinde2021automated}). They employ NLP techniques like: parts of speech(POS) tagging, dependency parsing, NER, and coreference resolution for candidate extraction. Nevertheless, we notice that mere linguistic clues are inadequate to capture all these stakeholders with their correct classes. For example,

\begin{quote}
    In the press conference, the oppositions criticises minister's comment and demanded his resignation.
\end{quote}

Here, \textit{opposition} and \textit{minister} both are geopolitical entities (GPE). But in political news scenario these two entities are classified as two different entity types [\textit{opposition}: \textbf{PoliticalParty:Opposition}; \textit{minister}: \textbf{Government}]. Some researchers also try to link entities with existing knowledge bases (e.g., Wikidata~\footnote{\url{https://www.wikidata.org/wiki/Wikidata}}, DBpedia~\footnote{\url{https://www.dbpedia.org/}}, YAGO~\footnote{\url{https://yago-knowledge.org/}}) as a way to employ external knowledge to recognize the correct entity types(~\cite{HACHEY2013130}). However, entity linking might be troublesome for stakeholder identification as most of the time exact proper name is not mentioned in the news reports. These stakeholders could be mentioned as proper nouns, nominal or pronominal mentions. Therefore, the task of stakeholder identification is not a trivial problem and has plausible importance in news analysis. 

In this work, we consider news articles based on four Indian government policies: Farmers’ Bill~\footnote{\url{https://en.wikipedia.org/wiki/2020_Indian_agriculture_acts}}, CAB~\footnote{\url{https://en.wikipedia.org/wiki/Citizenship_(Amendment)_Act,_2019}},  Demonetisation~\footnote{\url{https://en.wikipedia.org/wiki/2016_Indian_banknote_demonetisation}} and COVID management~\footnote{\url{https://en.wikipedia.org/wiki/Indian_government_response_to_the_COVID-19_pandemic}} published by several national and international news agencies and propose a generalized system for stakeholder extraction from these news articles. The task of stakeholder extraction consist of identification of stakeholder candidates and recognize their appropriate types. Each of these news-domain has a number of topic specific stakeholders that the reporters mention in the corresponding news documents. However, one silverline here is that each of these India specific news topics has some common set of stakeholders. To address the prespecified challenges we first create a well defined ontology to understand the concepts and coprehensive description regarding various stakeholders relevant to these topics and their inter-relations. We also annotate a small set of news articles where all the stakeholder candidates are labeled with proper stakeholder type that is used to evaluate the system performance. We employ our proposed model in these news articles, identify all the mentioned stakeholder phrases with corresponding classes, and report the visibility of each pre-specified stakeholder type per media outlet based on their mentions in the corresponding news reports. 


     
     


\section{Related Work}
Most of the studies in this field revolves around stakeholder identification from the political news coverage in two-party system of United States ~\cite{Covert2007MeasuringMB}~\cite{d2000media}.  Only a few works are dedicated to analyze the stakeholder bias in multiparty system~\cite{Hopmann2012PoliticalBI}.  In ~\cite{doi:10.1177/0093650215614364} authors analyze the salience of different stakeholders in the news coverage on Austrian presidential election campaign, 2013.
There are some papers which examined the distinct effect of the stakeholder references on generating media bias~\cite{Hallin2004ComparingMS}~\cite{Takens2010OldTF}~\cite{doi:10.1177/0093650215614364}. Works like ~\cite{cover_covid} and ~\cite{cover:climate} tried to measure the level of polarization and politicization on the respective COVID-19 and climate-change news based on the coverage of specific political parties and scientists.~\cite{gronberg2021extracting} build a system to identify the salient named entities from financial news articles. In contrast, our system deals with multiple stakeholder types relevant to various news topics and the outcome can be used for coverage bias identification.

\section{Overview and Problem Formulation}

Given a news article, we determine the visibility of a stakeholder type by the number of times the specific stakeholder candidates are referenced in that news article. Accordingly, the visibility score of a (stakeholder, media-house) pair is measured by the total number of times the stakeholder candidates are mentioned in the news articles published by that media house. Therefore, our objective here is to identify all the mentions of the pre-defined stakeholder candidates in the news articles.
Our experiment analyzes the news articles and finds some challenging issues regarding stakeholder extraction.

\begin{itemize}
    \item \textit{Selection:} All the named entities present in a news article are not considered potential stakeholders. We have to segregate those actors whose visibility, opinions, and claims would influence readers' judgment and perception of the issues and may intrigue bias in the news reporting.

\item \textit{Variation:} The same entity is referenced by more than one representation (e.g., full name, partial name, abbreviation, designation). It is crucial to identify all these coreferent mentions and map them with a single stakeholder class.

\item \textit{Ambiguity:} An entity mention could refer to different entities in different contexts. Furthermore, these mentions may belong to different stakeholder classes. So, it is essential to distinguish between these namesakes.

\item \textit{Grouping:} Group the entities of same concepts.
\item \textit{domain adaptation:} Generally, the stakeholders are domain-dependent. For instance, probable stakeholders in the entertainment domain may not be valid for the political news domain. Therefore, it is hard for a domain-specific model to transfer between other domains.

\end{itemize}

In this paper, we propose a model for stakeholder identification that addresses all these issues and extracts relevant stakeholder entities from the news articles generated from the real-time news streaming system. Furthermore, in order to ensure that all synonymous candidates are classified as the same stakeholder type, we must recognise all synonymous stakeholder mentions. The model identifies synonymous stakeholder entities within and across documents and sequentially constructs synonymous entity clusters (making prediction from left to right in the text). Eventually, these entity clusters are mapped to the potential stakeholder classes. Additionally we enable the model to identify new stakeholders in a completely unseen target news domain with a small number of labeled examples. In the following section, we first describe the stakeholder ontology, constructed based on the political news domain, then discuss the proposed method for synonymous stakeholder clustering and classification, and finally, check the model's effectiveness in other topics.

Initially, we have with us a list of news topics $T = [T_1, T_2,..., T_n]$ and a set of topic-specific stakeholder sets $S = \{S_1, S_2,..., S_n\}$, where topic $T_i$ contains pre-specified stakeholder set $S_i$. Meanwhile, given the news documents $D = [d_1, d_2,...]$ on topic $T_k$ generated by a news streaming system, our first task is to identify all the candidate stakeholder phrases $E= \{e_1, e_2, e_3,.., e_p\}$ present in each document $d_i$. Then, we aggregate all the synonymous entities across documents $D$ and generate a list of synonymous stakeholder clusters $L$. Eventually, we map each stakeholder cluster $L_j$ to correct stakeholder type $s_c$ from the pre-defined stakeholder type set $S_k$. More specifically, Our final output per document $d_i$ is a list of order pair: $(e_j, s_k)$ where, $e_j$ is the candidate stakeholder phrase and $s_k$ is the stakeholder Class. We store these order pairs in a dictionary $O^f$. During inference, first, we find the candidate phrase in the dictionary $O^f$ to determine the class. If the stakeholder is not in the dictionary, we try to find its synonymous stakeholder cluster in $L_j$ to decide its class.

\section{Stakeholder Ontology}

News articles on disputed topics contain multiple stakeholders with conflicting ideologies and opinions. Unlike most previous papers that are based on a two-party system~\cite{Covert2007MeasuringMB},~\cite{d2000media}, our work deal with multiple stakeholders relevant to the news topics. In this paper, we work with news articles based on four Indian government policies. India has a federal republic that constitutes a central authority and several states governed by elected representatives from multiple political parties~\footnote{\url{https://en.wikipedia.org/wiki/Federalism_in_India}}. India's pluralistic, multi-ethnic political and social stature makes these news articles an ideal case study for stakeholder analysis. We examine the news articles to identify the plausible candidate classes that are frequently mentioned in those stories and could be considered valid stakeholders. We focus on stakeholders whose visibility or salience may cause bias in a news publication. We devise an ontology to distinguish the stakeholder candidates based on their political ideology, social status, and topic-specific interests.

\begin{figure}[h]
  \centering
  \includegraphics[width=.99\linewidth]{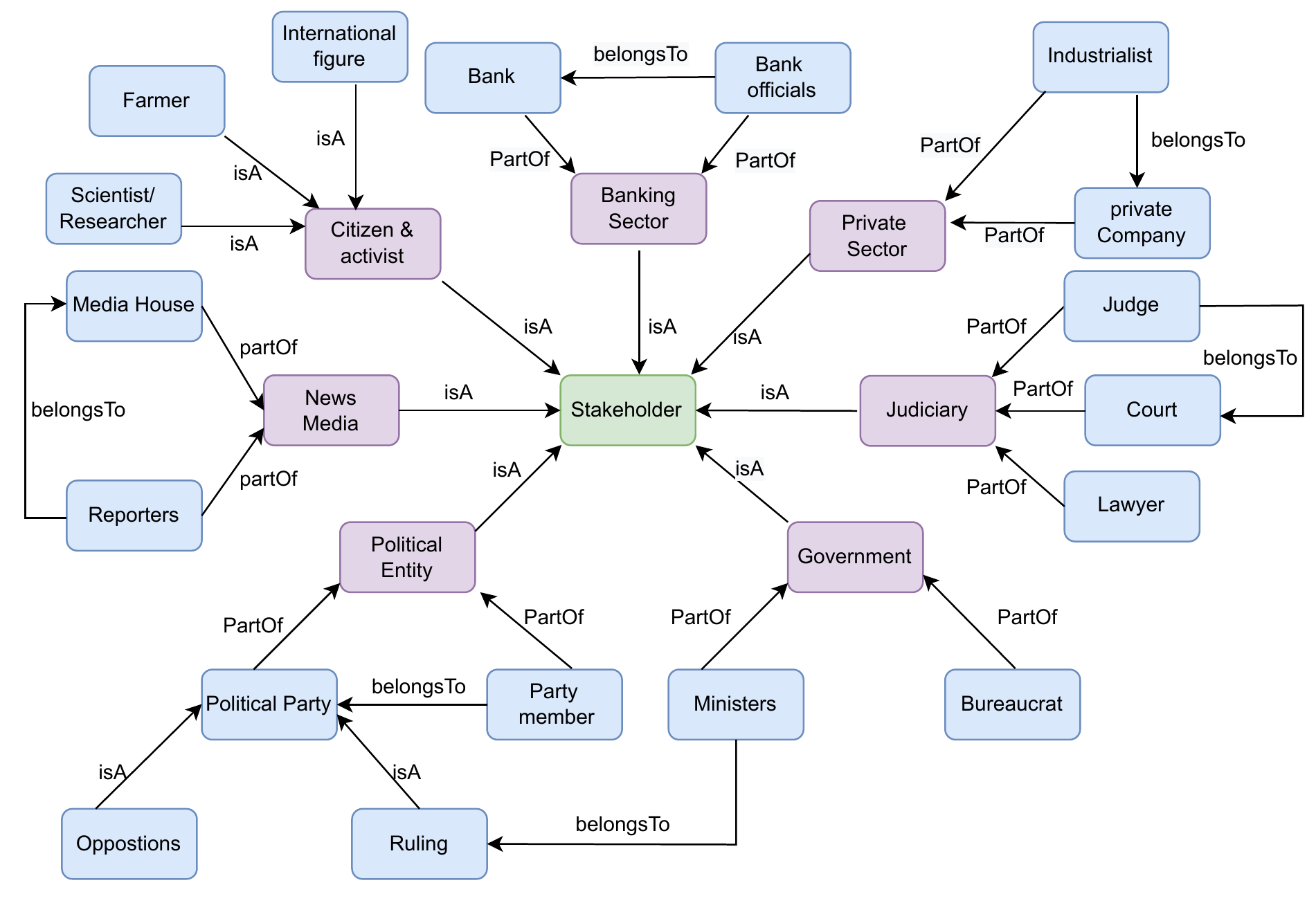}
  \caption{Stakeholder ontology graph}
  \label{fig:ontology1}
\end{figure}

The relevant stakeholder classes and their corresponding relations are depicted in Figure~\ref{fig:ontology1}. We define links between the ontology concepts by using the relations like: 1)isA, 2)belongsTo, and 3)partOf. In Indian multi-party system, political entities are an important stakeholders. The political entities are constitute of \textit{Political party} and ~\textit{party member}. The ~\textit{political parties} are further classified into ~\textit{ruling} and ~\textit{oppositions}. the ~\textit{Government} is another important stakeholder that is responsible for policy making. The ~\textit{government} consists of ~\textit{ministers} and ~\textit{bureaucrats}. However, the ~\textit{ministers} belong to ruling ~\textit{political parties}. ~\textit{Judicial authorities} are another important stakeholder that includes ~\textit{courts}, ~\textit{judges} and ~\textit{lawyers}. In the political news coverage, the media agencies play an important role and could be considered as potential stakeholders. Additionally, stakeholder type ~\textit{citizen \& Activist} are an integral part of any civilized society and their voices are equally important and influential. They can be further classified based on the topics and policies. Other prospective beneficiaries of the government policies are \textit{private sector} and \textit{banking sector} and could be considered as stakeholder based on specific news topic. All the relevant stakeholder classes, their sub-classes, components and inter-relations are reported in Figure~\ref{fig:ontology1}.  Most of the stakeholders are common in all four topics. However, based on the topic relevance, we conceive some topic-specific stakeholder classes as well.
Some of the topic-specific stakeholders are: \textit{Farmers} [topic: Farms' Law], \textit{Banking Sector}[topic: Demonetization], \textit{Private Company} [topic: Demonetization], \textit{Scientist/Researchers} [topic: Covid pandemic], \textit{International Figure} [topic: CAB bill]. In the news articles, these stakeholders are referenced by proper nouns or nominal mentions or pronominal mentions. Table~\ref{table:stakeholder_def} provides a brief description of some important stakeholder types with few candidate instances.  

For the sake of brevity in coverage analysis we select a subset of these stakeholders whose coverage or visibility may intrigue bias in news reporting. For instance, \emph{Government:ministers} and \textit{Political Party:ruling} are selected as the \textit{policy makers}. Media outlets which supports the government policies would favour the policy makers by giving them more coverage. Conversely, the visibility of \textit{Political party:oppositions} is equally important as the publication houses that criticise those policies would afford more salience to the opposition political parties. 
Table~\ref{table:stakeholder_types} represents the topic specific stakeholder types that are considered for coverage analysis.

\begin{table*}[ht]
\center\fontsize{6}{7.5}\selectfont
\begin{tabular}{cll}
\hline
Stakeholder Type           & \multicolumn{1}{c}{Definition}                     & \multicolumn{1}{c}{Examples}                                \\ \hline
Government     & Elected government act as policy makers                                      & Prime minister, Other cabinet ministers                     \\
Political Party:ruling     & Political parties in the power of central govt     & Political party members, spokespersons, MP, MLAs \\
Political party:opposition & Political parties acting as oppositions            & Political party members, spokespersons, MP, MLAs       \\
Judiciary                  & System of the court that interprete and apply laws & Supreme Court, High Courts, Chief Justice, Judges           \\
Govt. bureaucrat           & Non-elected government officials and executives    & Police and Army officials, IAS officers                     \\
Civic Society              & Consumers of the government policies               & Citizen, Activists, Public figures, NGOs                    \\

International Figure               &  Person or organization with international acceptance      & WHO~\footnote{url{https://www.who.int/}}, World Bank, Leaders of other countries                                      \\ 

News Editors               & News agencies who produce and circulate news       & Different media houses                                      \\ \hline
\end{tabular}
\caption{Description of some generic Stakeholders}
\label{table:stakeholder_def}
\end{table*}

\begin{table*}[!ht]
\center\fontsize{7}{9}\selectfont
\begin{tabular}{cl}
\hline
Topic          & \multicolumn{1}{c}{Stakeholders}                                                                                                                              \\ \hline
Farms' Law     & \begin{tabular}[c]{@{}l@{}}Government,  Opposition, Citizen/Activists, \textbf{Farmers}, \textbf{International-figure}\end{tabular}                                  \\ \hline
Demonetization & \begin{tabular}[c]{@{}l@{}}Government, Opposition, Citizen/Activists, \textbf{Banking Sector}, \textbf{Private Companies}\end{tabular}         \\ \hline
CAB Bill       & \begin{tabular}[c]{@{}l@{}}Government,  Opposition, Citizen/Activist, \textbf{International-figure}\end{tabular}     \\ \hline
Covid Control & \begin{tabular}[c]{@{}l@{}}Government, Opposition, Citizen/Activist\end{tabular}, \textbf{Scientist/Researchers}, \textbf{International-figure} \\ \hline
\end{tabular}
\caption{The Stakeholder considered for coverage analysis in each news topic. Topic-specific stakeholders are indicated in bold fonts.}
\label{table:stakeholder_types}
\end{table*}

\section{Model Framework}
\begin{figure}[!h]
  \centering
  \includegraphics[width=.95\linewidth]{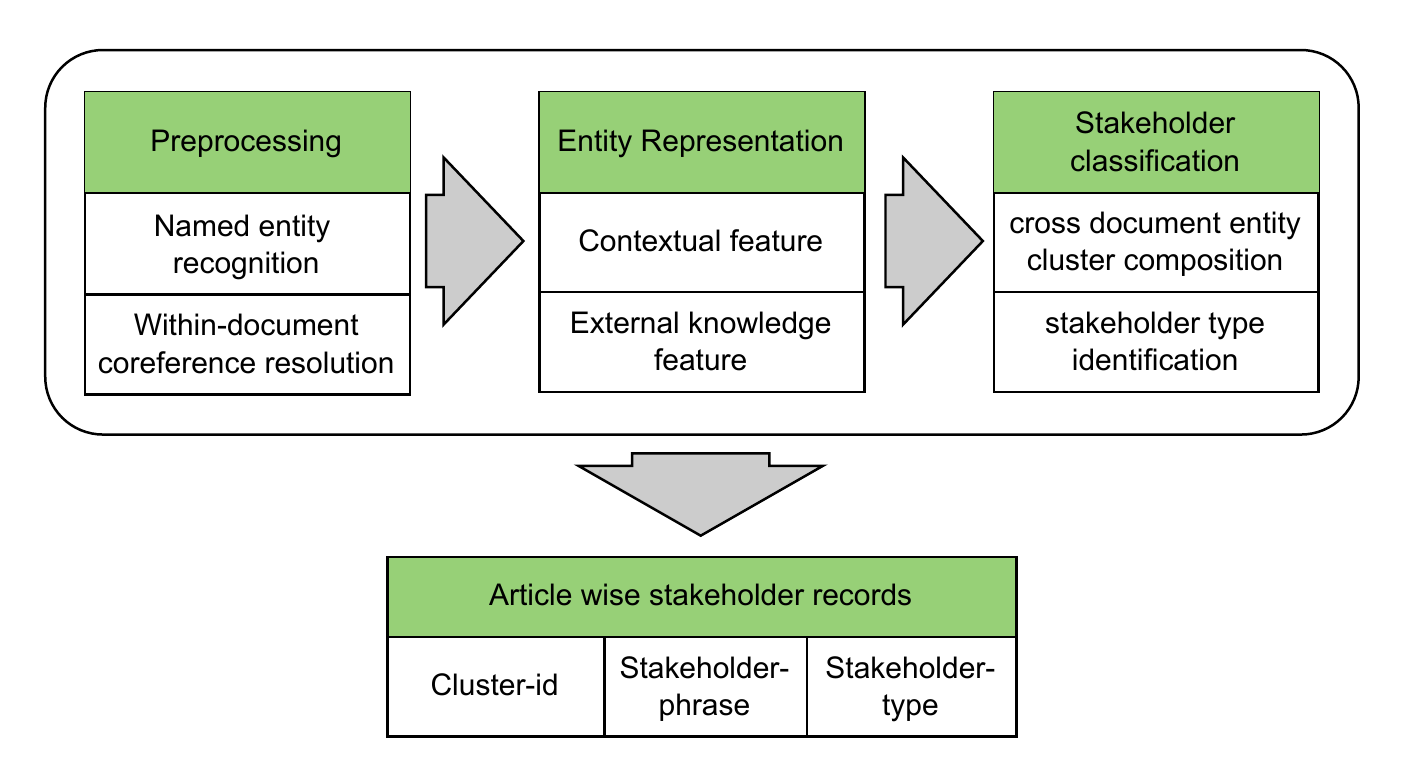}
  \caption{Stakeholder identification pipeline}
  \label{fig:pipeline}
\end{figure}

We now present the model that we propose for the task of stakeholder extraction from news articles. The proposed framework is able to extract all the mentions of stakeholder candidates, identify the synonymous stakeholder phrases in a cross-document scenario and reduce computational complexity. Recent advances in within-document entity coreference resolution demonstrate that sequential prediction (predicting coreferences from left to right in a text) achieves high performance with lower computational cost~\cite{lee-etal-2017-end}. Taking motivation from similar sequential coreference resolution approaches proposed by~\cite{Allaway2021SequentialCC}, we show that the technique can be extended to stakeholder extraction from cross-document news articles as well. The proposed system recognizes known entities from the previously made classification decisions stored in a stakeholder table, discovers new stakeholder-candidate phrases, and identifies the proper class of the stakeholder candidates based on our proposed candidate composition algorithm. Figure~\ref{fig:pipeline} gives a general workflow of our proposed system.

\subsection{Stakeholder phrase Recognition}

First, we aim to extract all the stakeholder phrases present in a news document. In this step we employ Spacy's~\footnote{\url{https://spacy.io/}} entity recogniser module~\footnote{\url{https://spacy.io/api/entityrecognizer/}} to identify all the entity mentions. However, all these entities (e.g.,\textit{place}, \textit{date}) are not considered as viable stakeholders. In our experiment, we consider \textit{Person} and \textit{Organization} type entities as potential stakeholder candidates. 



\subsection{Stakeholder phrase Representation: } To construct the representation for each entity mention $e$, we combine the following features: 

\begin{itemize}
    \item Contextual entity feature ($h_e$): In news articles, the person entities are often mentioned with their assigned positions, for example, ~\textit{Prime minister} Narendra Modi, ~\textit{Police commissioner} Rakesh Asthana, or ~\textit{Opposition leader} Rahul Gandhi. The phrases \textit{prime minister, police commissioner, and opposition leader} play a vital role in determining the proper stakeholder type of the candidate entity phrases and are incorporated into the entity phrase representation.
    The contextual feature of the entity generates from the embeddings of the start and end sub-word tokens of the entity span containing the head-word and the modifier information. We get the entity spans by applying semantic role labelling~\footnote{\url{https://pypi.org/project/transformer-srl/}} on the sentences.
    
    \item Global entity feature ($h_{KB}$): We study that mere sentential information is not enough to identify all the stakeholder classes. Sometimes domain information in the form of background knowledge is essential to identify the proper stakeholder types. Therefore, we attempt to link the extracted stakeholder phrases with their corresponding Wikipedia page and collect initial introductory sentences from that page to congregate more domain information about them. We use Wikipedia\footnote{\url{https://pypi.org/project/wikipedia/}}, a Python library that allows us to easily access and parse Wikipedia data. However, the system fails to identify the proper Wikipedia page for some stakeholder phrases. To resolve the issue, we manually fetch a number of India-related Wikipedia pages that depicts Indian politicians, public figures, Indian government policies, and recent socio-economical and political events that happened in recent years and collect a considerable amount of India-specific textual content. Later, we apply semantic role labelling~\footnote{\url{https://pypi.org/project/transformer-srl/}} on those sentences and accumulate a list of (subject, predicate, object) triplets. The \textit{subject} and \textit{object} are the argument phrases containing some entity information that could be matched with the stakeholder candidate phrases of our interest. We collect those triplets and embed them in order to obtain the stakeholder phrase representations.
    
\end{itemize}
 We aggregation these feature vectors and produce the individual entity embedding that combines both contextual and external feature embedding. For aggregation we use mean-pooling.

     $$h_{x} = h_{e} + h_{KB}$$
Here, $h_{x}$ indicates the embedding of entity mention $e$, 
 $h_{e}$ is the embedding of the entity span containing the head-word and the modifier information of entity $e$ . The description about the entity collected from Wikipedia page is embedded and represented by $h_{KB}$.



\subsection{Example based Stakeholder Identification}
In general, the goal of example-based stakeholder identification is to perform stakeholder classification using a few examples as support-set for any stakeholder type, including previously unknown stakeholder types. For example, given a sample sentence \say{Honourable Prime Minister Narendra Modi delivered the inauguration speech in the event}  containing stakeholder type \textit{Govt:elected},  where \textit{Narendra Modi} is the candidate phrase, our expectation is to recognize the entity phrase \textit{Arun Jately} as of type \textit{Govt:elected} in the query sentence \say{Finance minister Arun Jaitley met the business personalities in that event.} It shows a single support example for each stakeholder type scenario. The real-world scenarios considered in this paper involve various stakeholder types, with a few samples for each.

\subsection{Sequential cross document candidate clustering: }

In real-world scenarios, a specific topic is discussed in several articles; hence similar stakeholder candidates are mentioned in those collections of articles~\cite{lee-etal-2017-end}. To identify the synonymous stakeholders, we have to expand the discourse and resolve the entities across the documents~\cite{Allaway2021SequentialCC}. Our objective is to identify the synonymous candidates across documents and keep them in a single cluster. 
We apply a sequential model that iterates through the list of documents and predicts coreference links between entities in the current document and existing candidate clusters computed across all the preceding documents. Specifically, for every mention, a coreference decision is made based on the current state of coreference clusters rather than a set of individual mentions. In this way, the model can make decisions based on information about the mentions that are currently in a cluster. Let $L_f$ be the set of coreference clusters and $L_f = {P_1, P_2, ..., P_n}$ where each $P_i$ is a set of synonymous entity mentions across documents. $Y$ is a list of stakeholder labels where $Y_i$ holds the stakeholder label of $P_i$. For instance, $P_i = [ (e_1,D_3), (e_5,D_7), (e_p,D_q)]$ where $e_p$ is the entity phrase and $D_q$ is the document identifier. Here, $(e_1,D_3)$ indicates entity $e_1$ presides in document $D_3$ and $e_1$, $e_5$ and $e_p$ are synonymous entities with stakeholder type $Y_i$. Our task is to assign a new stakeholder phrase $x_q$ to one of these clusters and determine its correct stakeholder-type. We compute the similarity between the query entity $x_q$ and candidate clusters $L_f$ using some similarity (Cosine similarity) metrics to populate these clusters. To allow the model to predict singleton mentions, we check if the maximum similarity score is less than some predefined threshold. In the affirmative case, we create a new singleton cluster instead of assigning $x_i$ to an already existing cluster. We sequentially populate the synonymous entity sets and update entity cluster set $L_f$ in each iteration. Note that $L_f$ consists of gold entities mentioned as seed clusters. So each cluster initially is provided with some entities whose labels ($Y$) have been pre-determined.

\begin{algorithm}
\centering
\begin{minipage}{\columnwidth}\scriptsize{}{}\selectfont
\caption{Sequential cross document candidate clustering algorithm}\label{alg:cap}
     \textbf{Input} \\
     $L^0$ \Comment{initial seed dictionary with \textit{key:}EntityId, \textit{value:} seed synonymous candidates with their labels }\\
 $E$ \Comment {List of extracted entities from the news articles }\\
$T$ \Comment {Threshold similarity score}\\
    \textbf{Output} 
     $L^f$, $Y$  \Comment {Final set of candidate clusters with labels}
\begin{algorithmic}[1]

\State $L^f \gets L^0$
\For {each $e \in E$}
\If{$e \in L^h$}\Comment{where, $h \in [1, |L^f|]$}
\State $Label(e) \gets Y_h$
\Else
\State $h_e \gets Embedding(e)$
\State $h_k \gets Embedding(L^f_k)$ \Comment{where, $k \in [1, |L^f|]$}
\State $m \gets argmax\{ComputeSimilarScore(h_e, h_k)\}$ \Comment{$\forall k \in [1, |L^f|]$}
\State $SimScore \gets ComputeSimilarScore(h_e, h_m)$
\If{$SimScore > T$}
\State $L^f_m \gets L^f_m \cup CandidateHead(e)$
\State $Label(e) \gets Y_m$
\Else
\State $L^f \gets L^f \cup \{e\}$ \Comment{candidate inserted in a new candidate cluster}
\State $Label(e) \gets Y_m$
\State $Y_{|Y^f|+1} \gets Y_m$
\EndIf
\EndIf
\EndFor
\State \textbf{return} $L^f , Y$
\end{algorithmic}
 \end{minipage}
\end{algorithm}

\section{Experiment}

\subsection{Dataset}

Although stakeholder mentions are regarded as a crucial factor in news analysis, we are unable to locate any reasonably sized labeled corpus that suits our research, which is why existing works rely on dictionary-based stakeholder identification. We use GDELT~\footnote{\url{https://blog.gdeltproject.org/the-datasets-of-gdelt-as-of-february-2016/}} as our news source to obtain topic-oriented news articles. The GDELT dataset, which contains geo-located events published in news reports from around the world, provides daily coverage of events found in news reports published daily. We have collected data relevant to
our topics from GDELT Dataset. Since the data contained only URLs of the articles, we scraped
the actual content from these URLs provided in the GDELT GKG table\footnote{\url{https://blog.gdeltproject.org/}} using various scraping tools. 

\paragraph{News Domain}
Initially, we scrapped around 83k India-oriented news articles from 42 media sources. However, in our experiments, we consider news articles on four Indian Govt policies:
1) Farmers’ Law, 2) Demonetization, 3) Citizenship Amendment Bill (CAB), 4) COVID pandemic management. Therefore we apply semi-supervised topic modeling to get topic-oriented news articles. We also annotate potential stakeholder candidates with their types in some of these extracted articles. The remaining unlabeled news documents are used for stakeholder coverage analysis. The following table reports the data statistics extracted and used in our experiment. 

\begin{table}[!ht]
\centering
\resizebox{\columnwidth}{!}{%
\begin{tabular}{|c|c|cc|}
\hline
\multirow{2}{*}{\textbf{Topic}} & \multirow{2}{*}{\textbf{Publish Time Range}} & \multicolumn{2}{c|}{\textbf{Dataset count}} \\ \cline{3-4} 
                                &                                              & \multicolumn{1}{c|}{Unlabeled}   & Labeled  \\ \hline
Farm Law                        & 2020-8-10 to 2021-11-30                     & \multicolumn{1}{c|}{9854}       & 50       \\ \hline
Demonetization                  & 2016-11-08 to 2021-11-23                     & \multicolumn{1}{c|}{5301}        & 27       \\ \hline
CAB Bill                        & 2019-09-08 to 2020-04-25                     & \multicolumn{1}{c|}{3003}        & 22       \\ \hline
Covid management                & 2020-03-30 to 2022-03-01                     & \multicolumn{1}{c|}{27987}       & 20       \\ \hline
\end{tabular}%
}
\end{table}

\paragraph{News Sources}
For news sources, we select media agencies with valid references in ~\url{media-rank.com}. These media houses could be classified as foreign and Indian media houses. During the content analysis, we report the variation of stakeholders' coverage by different media houses. The news outlets whose news stories we used in our experiments are listed in the table below.

\subsection{Domain Adaptive pre-training}
There is a domain mismatch because BERT was trained on BooksCorpus and Wikipedia, and our experiments are based on news articles~\cite{devlin-etal-2019-bert}. Furthermore, using a domain corpus for pre-training aids in addressing the data scarcity issue of the entities~\cite{ma-etal-2020-resource}. As a result, prior to using our models, we fine-tune BERT using a pool of news articles (50k documents) with masked language model (MLM) loss, as this is effective for domain transfer~\cite{gururangan-etal-2020-dont}.

\subsection{System Performance}
In this section, we present the experimental results that demonstrate the generality of our model, the ease with which it can be applied, and the effectiveness of the proposed method.

\begin{table}[!ht]
\centering
\resizebox{\columnwidth}{!}{%
\begin{tabular}{cccc}
\hline
Stakeholder Type          & Precision & Recall & F-Score \\ \hline
CentralGovt:Elected       & 79        & 82     & 80      \\
Judiciary                 & 86        & 83    & 84      \\
Bureaucrat                & 75        & 79     & 77     \\
PoliticalParty:Ruling      & 76        & 73     & 74      \\
PoliticalParty:Opposition & 79        & 75     & 77      \\
International Figure      & 71        & 76     & 73      \\
Bank Sector               & 82        & 86     & 84      \\
Citizen/Activist          & 85        & 74     & 79      \\
\textbf{Macro-Fscore}              & -         & -      & 79      \\ \hline
\end{tabular}%
}
\end{table}

\paragraph{Time Complexity}

Let $D$ be a collection of documents containing $M$ stakeholder mentions that form $C$ coreference clusters. Our proposed sequential algorithm, which computes similarity scores between a specific mention and all antecedents, must compute $CM$ scores. In contrast, in a typical graph-based model, computing $\approx M^2$ scores between all pairs of mentions in all documents is always required. Our model is more efficient in practise because $C \ll M$. This efficiency boost is even more important in real-world streaming scenarios.

\section{Stakeholder coverage}
\begin{figure*}[ht] 
  \begin{subfigure}[b]{0.5\linewidth}
    \centering
    \includegraphics[width=0.95\linewidth]{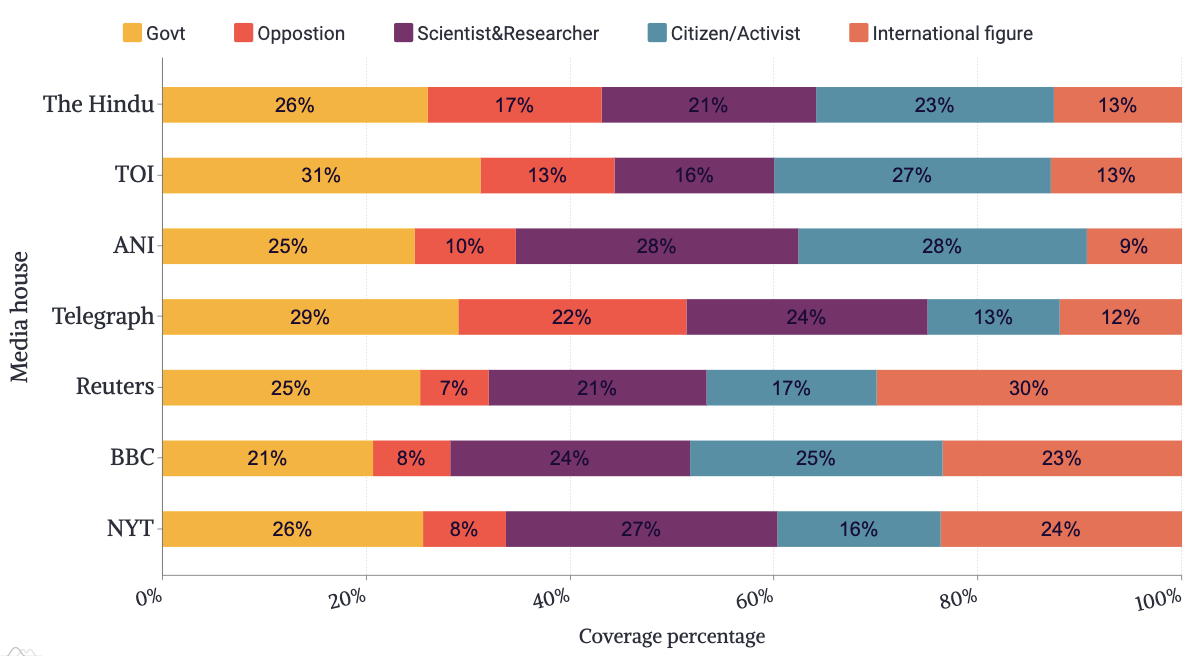} 
    \caption{Covid management} 
    \label{stat:a} 
    \vspace{4ex}
  \end{subfigure}
  \begin{subfigure}[b]{0.5\linewidth}
    \centering
    \includegraphics[width=0.9\linewidth]{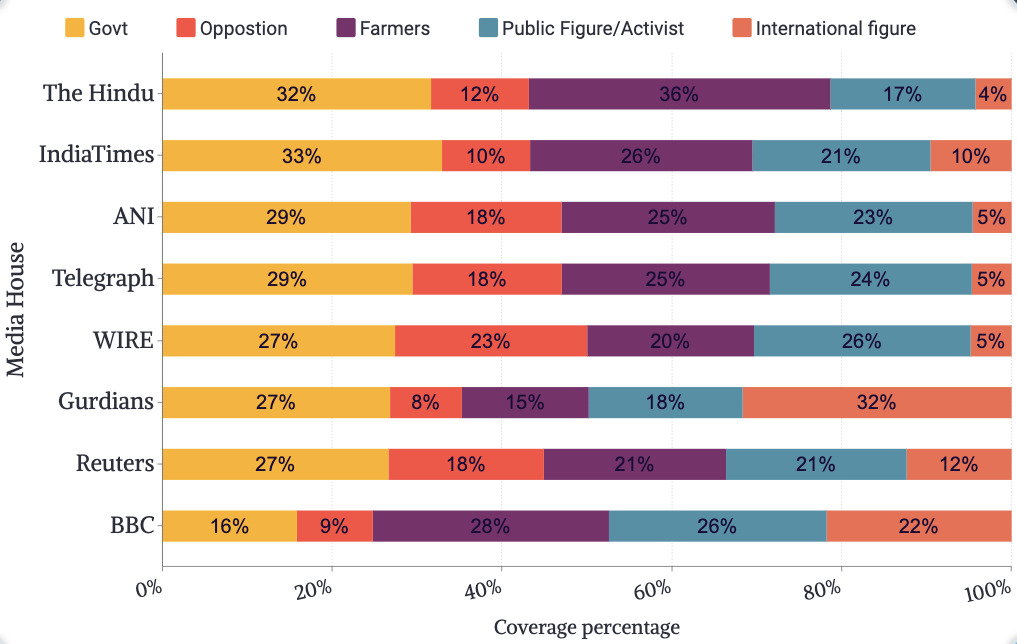} 
    \caption{Farm' Bill} 
    \label{stat:b} 
    \vspace{4ex}
  \end{subfigure} 
  \begin{subfigure}[b]{0.5\linewidth}
    \centering
    \includegraphics[width=0.95\linewidth]{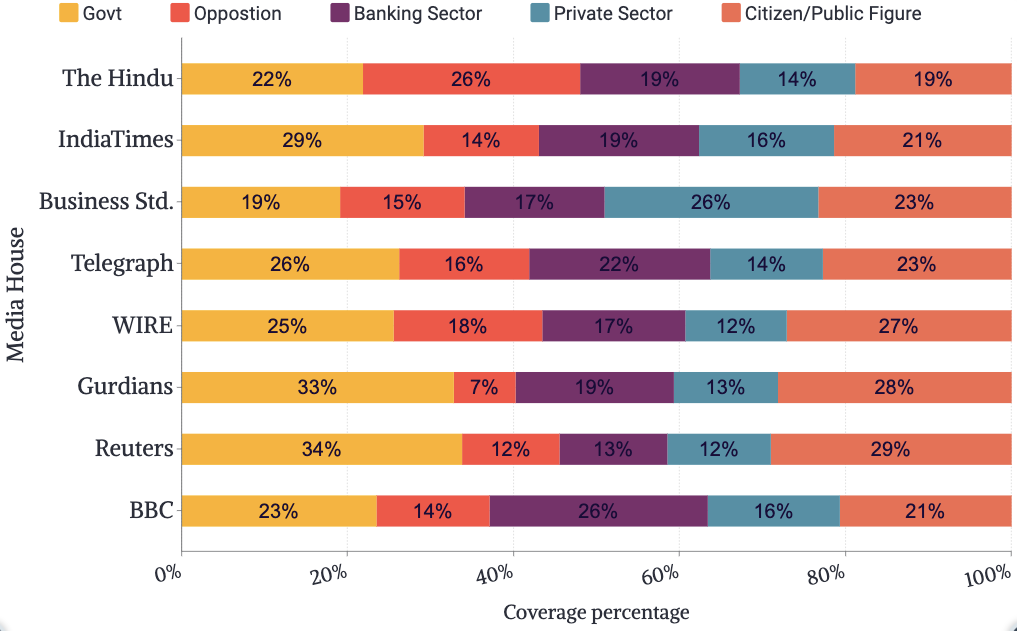} 
    \caption{Demonetization} 
    \label{stat:c} 
  \end{subfigure}
  \begin{subfigure}[b]{0.5\linewidth}
    \centering
    \includegraphics[width=0.95\linewidth]{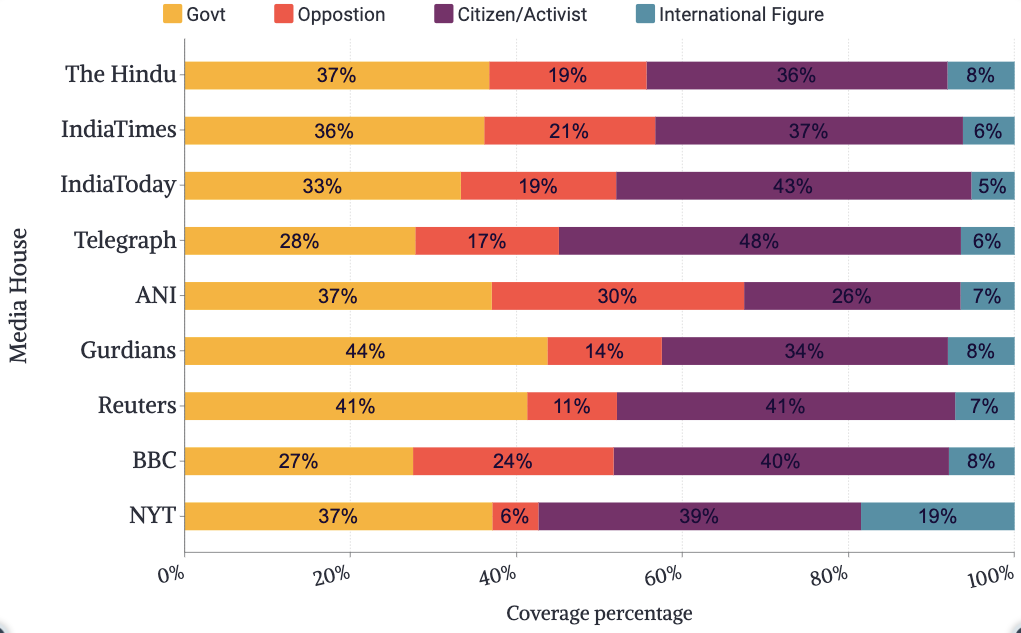} 
    \caption{CAB Bill} 
    \label{stat:d} 
  \end{subfigure} 
  \caption{Difference in coverage among (stakeholders, media house) pairs per news topics}
  \label{coverage:stat} 
\end{figure*}

In this section, we present the experimental results that demonstrate the generality of our model, the ease with which it can be applied, and the efficacy of the proposed visualizations. We examine stakeholder coverage in extracted news articles based on four predefined topics. In this paper, we define the coverage or visibility of that stakeholder class by the number of times a stakeholder candidate is mentioned in a specific news article. Eventually, the total number of times the stakeholder candidates are referenced in the articles published by a media house determines the overall coverage or visibility for each (stakeholder, media house) pair. We use the same sequential candidate clustering algorithm to identify and classify all stakeholder candidates in the remaining unannotated news articles. Figure~\ref{coverage:stat} shows the percentage of coverage for each stakeholder mentioned in the corresponding media houses' reporting.




  

\subsection{Coverage Analysis}
In this section, we examine the pattern of stakeholder coverage in various media publications. We report the media-house specific coverage score of various stakeholders relevant to topic \textit{Covid management} in Figure~\ref{coverage:stat}(a).  For coverage analysis, we consider both national and foreign news agencies. Most of the news agencies offer more coverage to the \textit{Government}, \textit{Scientist \& Researchers} and \textit{Citizen/Activists}. However, the foreign news agencies accord comparatively more coverage to the International figures. Most of the publication (except \textit{Telegraph}) gives less amount of coverage to \textit{Oppositions}. The coverage scores of different stakeholders related to the topic: \textit{Farm' Bill} are shown in Figure. ~\ref{coverage:stat}(c). In topic ~\textit{Farm Bill}, the most covered stakeholders are ~\textit{Government}, ~\textit{Farmers} and \textit{public figure/ activist}. Even, \textit{The Hindu} gives more coverage to \textit{Farmers} than \textit{Government}. However, \textit{International figure} is a dominant stakeholder in foreign media publications. Surprisingly, in topic \textit{Demonetization}, \textit{Government} and \textit{Citizen/public figure} share similar coverage in most of the publication. \textit{Banking sector} and \textit{Private sector} also obtain substantial coverage.
In case of \textit{CAB bill}, the dominant stakeholders are \textit{Government } and \textit{Citizen/ activist}. In some cases, \textit{Citizen/ activists} receive more coverage than \textit{government} entities.

\section{Conclusion}
It is important to recognize our research's limitations, which are limited to empirical measures of coverage of different stakeholders in the news media. Our goal of this work is not to determine whether the media is biased in favor of one party. Our entire analysis is based on a collection of news articles accumulated over a specific time. The proposed model is also useful for identifying stakeholders in other news domains with minimum domain-specific information. In the future, we will thoroughly analyze the robustness of our system and focus on designing an efficient domain-agnostic stakeholder classifier. Our present work can also be extended to analyze inherent bias in media reporting based on the coverage of the stakeholders.


\begin{quote}
\begin{small}
\bibliography{aaai23}
\end{small}
\end{quote}

\end{document}